 \documentclass[pmlr,twocolumn,10pt]{jmlr} 

\usepackage[T1]{fontenc}
\usepackage{multirow}
\usepackage{booktabs}
\usepackage{verbatim}
\usepackage{stfloats}
\usepackage{enumitem}
\usepackage{siunitx}
\usepackage[leftmargin=1.5em]{quoting}

\theorembodyfont{\upshape}
\theoremheaderfont{\scshape}
\theorempostheader{:}
\theoremsep{\newline}

\definecolor{limegreen}{rgb}{0.2, 0.8, 0.2}

\jmlrvolume{}
\jmlryear{}
\jmlrsubmitted{}
\jmlrpublished{}
\jmlrworkshop{} 

\title[Retrieving Evidence from EHRs with LLMs: \\ Possibilities and Challenges]{Retrieving Evidence from EHRs with LLMs:\\ Possibilities and Challenges}

\author{%
\Name{Hiba Ahsan}\Email{ahsan.hi@northeastern.edu}\\
\addr Northeastern University, Boston, MA
\AND
\Name{Denis Jered McInerney}\Email{mcinerney.de@northeastern.edu}\\
\addr Northeastern University, Boston, MA
\AND
\Name{Jisoo Kim} \Email{jkim@bwh.harvard.edu}\\
\addr Brigham and Women’s Hospital, Boston, MA
\AND
\Name{Christopher Potter} \Email{cpotter3@bwh.harvard.edu}\\
\addr Brigham and Women’s Hospital, Boston, MA
\AND
\Name{Geoffrey Young} \Email{gsyoung@bwh.harvard.edu}\\
\addr Brigham and Women’s Hospital, Boston, MA
\AND
\Name{Silvio Amir} \Email{s.amir@northeastern.edu}\\
\addr Northeastern University, Boston, MA
\AND
\Name{Byron C. Wallace} \Email{b.wallace@northeastern.edu}\\
\addr Northeastern University, Boston, MA
}

\begin{document}

\maketitle

\begin{abstract}
Unstructured data in Electronic Health Records (EHRs) often contains critical information---complementary to imaging---that could inform radiologists' diagnoses. 
But the large volume of notes often associated with patients together with time constraints renders manually identifying relevant evidence practically infeasible. In this work we propose and evaluate a zero-shot strategy for using LLMs as a mechanism to efficiently retrieve and summarize unstructured evidence in patient EHR relevant to a given query.  
Our method entails tasking an LLM to infer whether a patient has, or is at risk of, a particular condition on the basis of associated notes; if so, we ask the model to summarize the supporting evidence. 
Under expert evaluation, we find that this LLM-based approach provides outputs consistently preferred to a pre-LLM information retrieval baseline. 
Manual evaluation is expensive, so we also propose and validate a method using an LLM to evaluate (other) LLM outputs for this task, allowing us to scale up evaluation. 
Our findings indicate the promise of LLMs as interfaces to EHR, but also highlight the outstanding challenge posed by ``hallucinations''. In this setting, however, we show that model confidence in outputs strongly correlates with faithful summaries, offering a practical means to limit confabulations.
\end{abstract}

\paragraph*{Data and Code Availability}
We describe the data used for evaluation in \S3. Briefly, we evaluate our approach using two datasets: (1) MIMIC-III dataset, \citep{johnson2016mimic}, which is available on PhysioNet \citep{johnson2016physionet}; and (2) EHR notes of patients admitted to the Emergency Room of Brigham and Women's Hospital (BWH) in Boston, MA, USA,  between 2010 and 2015. 
Our code and data are available at \url{https://github.com/hibaahsan/chil_diagnosis_evidence/}.

\paragraph*{Institutional Review Board (IRB)}
This retrospective medical records research was approved by the Mass General Brigham (MGB) IRB with a waiver of requirement for informed consent.

\section{Introduction}
\label{section:intro}

\begin{figure*}
    \centering
    \includegraphics[scale=0.55]{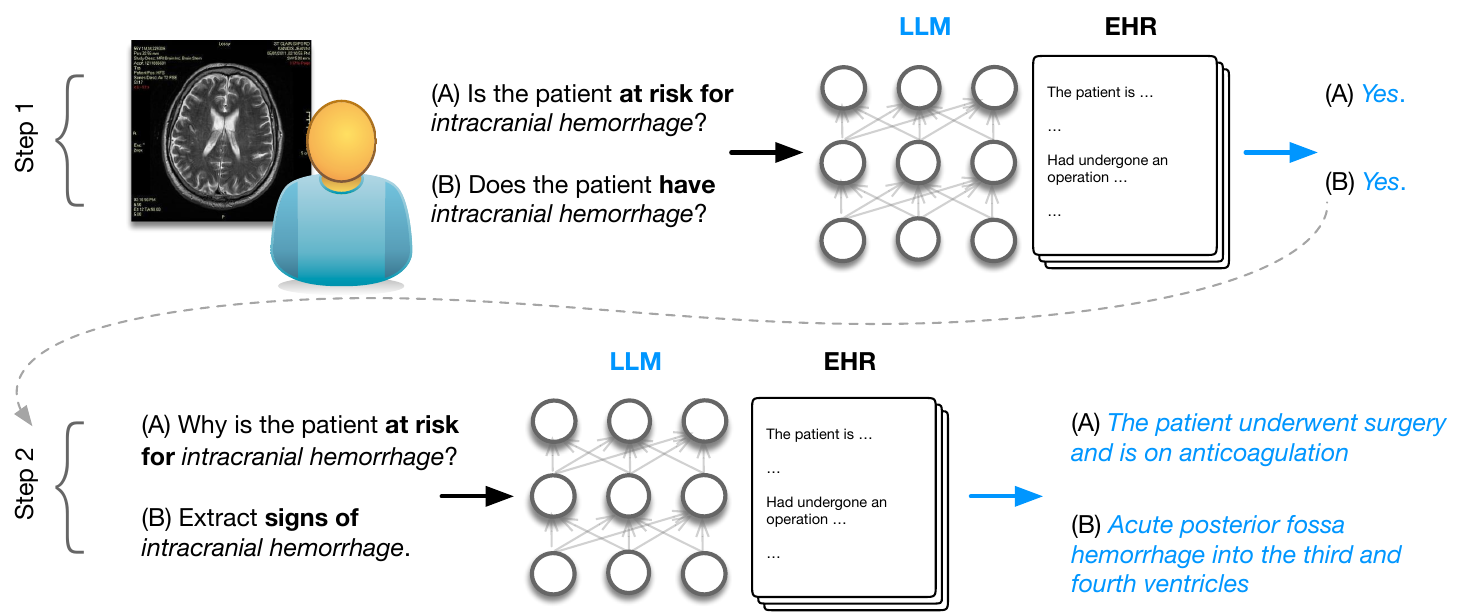}
    \caption{Proposed prompting strategy to identify and summarize evidence relevant to a given query diagnosis using LLMs. We first ask if the patient has (or is at risk of) a condition, then elicit a summary of supporting evidence if so.}
    \label{fig:ehr-flow}
\end{figure*}

We consider using 
LLMs as interfaces to unstructured data (notes) in patient Electronic Health Records (EHRs), ultimately to aid radiologists performing imaging diagnosis. 
The motivation is that unstructured evidence within EHR may support (or render less likely) particular diagnostic hypotheses radiologists come to based on imaging, but time constraints---combined with the often lengthy records associated with individual patients---make manually finding and drawing upon such evidence practically infeasible. 
Consequently, radiologists often perform diagnosis with comparatively little knowledge of patient history. 

LLMs offer a flexible mechanism to interface with unstructured EHR data, e.g., recent work has shown that LLMs can capably perform \emph{zero-shot} information extraction from clinical notes \citep{agrawal2022large,jered-emnlp-23}. 
In this work {\bf we propose and evaluate an approach using LLMs to extract evidence from EHR notes to aid diagnosis}.  
We envision a clinician providing an initial suspected diagnosis as a query; the LLM should then confirm whether there is unstructured (textual) evidence in the patient record that might support this diagnosis, and---if so---summarize this for the clinician (Figure \ref{fig:ehr-flow}).

LLMs provide an attractive mechanism to permit such interactions given their established dexterity working with unstructured text, and their flexibility. 
Critically, they permit general question answering (e.g., ``Is this patient at risk of \emph{Atrial fibrillation}?'') and can summarize supporting evidence. But with this flexibility comes challenges: Skillful as they are, LLMs are prone to ``hallucinating'' content \citep{azamfirei2023large,zhang2023language}, which is particularly concerning in healthcare.  

We conduct an empirical evaluation with practicing radiologists to assess the use of LLMs as diagnostic aids. 
Our results show that LLMs are more capable than a representative ``traditional'' (pre-LLM) information retrieval system at surfacing and summarizing evidence relevant to a given diagnosis. 
However, manual evaluation by domain experts does not scale. 
Therefore, we propose and assess an automated evaluation approach using LLMs. 
Given a piece of evidence, we enlist an evaluator LLM to: (i) Extract the conditions stated as risk factors (or signs) in this snippet; (ii) Confirm the presence of each condition in the note independently; and then (iii) Validate whether each condition is a risk factor (or sign) of the query diagnosis. 
We find that this automated assessment strategy correlates with expert evaluations, and therefore use it to scale up our evaluation. 

Our work shows the potential of LLMs as interfaces to EHRs, but also highlights challenges inherent to their use. 
How can we know that a generated summary of supporting evidence faithfully reflects an underlying patient record? 
We highlight troubling examples where the LLM fabricates plausible patient history that \emph{would} support a condition of interest. 
At best this frustrates the provider (who must read through the record carefully to ascertain if there is in fact such evidence), and at worst it is dangerous. 
However, we find that model confidence in generations strongly correlates with accuracy in this domain, which mitigates this issue. 

Our contributions are as follows. 
(1) We introduce an approach in which we task an LLM to infer patient risk of a given condition, and to produce a conditional summary of supporting evidence if so. 
We enlist experts to manually evaluate outputs from two LLMs---Flan-T5 XXL \citep{chung2022scaling} and Mistral-Instruct \citep{jiang2023mistral}---and find they both outperform a representative baseline evidence retrieval approach.
(2) We introduce a method to automate evaluation of retrieved evidence via an LLM, and show this enjoys good correlation with expert annotations. Larger scale evaluation using this approach confirms the advantage of LLMs over traditional methods. 
(3) We highlight examples that illustrate the issue of hallucinated content in this context, and report results indicating that LLM confidence may be sufficient to avoid this. 

\section{Retrieving and summarizing evidence with LLMs}
\label{section:approach} 

For a given query ($\equiv$ condition), we attempt to retrieve two distinct types of evidence from patient history: (A) snippets that indicate a patient \emph{may be at risk} of developing the condition in the future, and; (B) those that suggest the patient \emph{currently has} the condition. 
For example, a patient on anticoagulants after a recent posterior fossa surgery may be at risk of an intracranial hemorrhage, but not experiencing one currently. 
By contrast, observing acute posterior fossa hemorrhage indicates the patient most likely has intracranial hemorrhage.

Extracting evidence for \textit{risk} informs clinicians about occurrences in the patient's history (e.g., procedures, diagnoses) that make them more vulnerable to the condition. 
Extracting evidence for \textit{signs} of a condition serves two purposes. 
Those that occur in the patient's immediate history indicate that they likely have the condition; those that occur earlier indicate the patient (may) have a history of the condition, which is also important. 

We consider openly available ``medium-scale'' models, including Flan-T5 XXL \citep{chung2022scaling} and Mistral-Instruct \citep{jiang2023mistral} as representative LLMs (11.3B and 7B parameters, respectively). 
While larger, proprietary models may offer superior results, we wanted to use an accessible LLM to ensure reproducibility.
Moreover, protections for patient privacy mandated by the Health Insurance Portability and Accountability Act (HIPAA), and our institutional policy on use of LLM restrict us to using models that can be deployed ``in-house'', precluding hosted variants (e.g., those provided by OpenAI).

\paragraph{Zero-shot sequential prompting}

We adopt a sequential prompting approach to find and summarize evidence relevant to a query. 
We first ask the LLM whether a given note indicates that the corresponding patient is at risk for or has a given query diagnosis---prompting the LLM for a binary decision about this. If the answer is `Yes', we prompt the model to provide support for this response.

More specifically, to query whether the patient is at risk for the given diagnosis, we use the prompts below for Flan-T5 and Mistral-Instruct.

\begin{quote}
    \noindent Read the following clinical note of a patient: \texttt{[NOTE]}.
    
    \noindent Question: Is the patient at risk of \texttt{[DIAGNOSIS]}?\\Choice -Yes -No.
    
    \noindent Answer: 
\end{quote}

\noindent To elicit supporting evidence from the model for these risk predictions, we use the following prompt for Flan-T5.

\begin{quote}
    \noindent Read the following clinical note of a patient: \texttt{[NOTE]}.
    
    \noindent Based on the note, why is the patient at risk of \texttt{[DIAGNOSIS]}? 
    
     \noindent Answer step by step: 

\end{quote}

\noindent For Mistral-Instruct, we found that CoT prompting yielded very lengthy responses. We therefore instead used the following prompt:

\begin{quote}
    \noindent Read the following clinical note of a patient: \texttt{[NOTE]}.
    
    \noindent Based on the note, why is the patient at risk of \texttt{[DIAGNOSIS]}? Be concise.

    \noindent Answer:
\end{quote}

Similarly, to query whether the patient \emph{has} a given diagnosis, we ask ``Question: Does the patient have \texttt{[DIAGNOSIS]}?'' (asking for a binary response).
And then to obtain evidence supporting this assessment (in the case of a positive response), we prompt with: ``Question: Extract signs of \texttt{[DIAGNOSIS]} from the note.''. 
In the above prompts, \texttt{[NOTE]} denotes a patient note, and \texttt{[DIAGNOSIS]} a potential diagnosis for which we would like to retrieve supporting evidence. 
We then combine and present the result for the two types of evidence (risks and signs) to the end user.

\paragraph{Why not a single prompt?} It might seem more intuitive to simply ask the model to answer `Yes' or `No' \emph{and explain its reasoning} in a single prompt. 
However, we found that this strategy yielded many false positives for both Flan-T5 and Mistral-Instruct. 
To quantify this, we randomly sampled $40$ notes and used a single prompt to find evidence for conditions that the patient did \textit{not} have. 
The single prompt produced `No' for only $7.5\%$ (Flan-T5) and $27.9\%$ (Mistral-Instruct) of the notes. 
By contrast, sequential prompting yielded `No' all $40$ times for both models. 
We provide more details in  \S\ref{app:single-prompt}. 
We also experimented with a single few-shot prompt to extract evidence (\S \ref{app:few-shot}), but preliminary results were not promising so we did not pursue this further.

\paragraph{A retrieval baseline (CBERT)} As a point of comparison for unsupervised evidence extraction (with pre-LLM methods), we use a simple ranking approach using neural embeddings.\footnote{Other, even simpler, baselines are a possibility (e.g., BM25, TF-IDF), but the expensive expert time required for annotations limited our ability to evaluate additional baselines.}
Specifically, given a query  \texttt{[DIAGNOSIS]}, we retrieve associated \texttt{[RISK FACTORS]} using GPT-3.5 and generate an embedding $e_\textrm{rf}$ of the sentence: `Risk factors of \texttt{[DIAGNOSIS]} include \texttt{[RISK FACTORS]}' using ClinicalBERT \citep{alsentzer-etal-2019-publicly}.\footnote{Note that this does not entail passing any sensitive data to OpenAI; we send only a condition name.} 

Table \ref{tab:risk_factors} shows examples of risk factors provided by GPT-3.5.
The intuition is to generate $n$-grams that are likely to indicate risk of the corresponding diagnosis so that we can match these against notes in EHR. 
Then, for a patient and \texttt{[DIAGNOSIS]}, we retrieve the top $20$ sentences in the patient notes most similar to $e_\textrm{rf}$.
One downside of such a retrieval-based approach is the need to pre-specify the number of evidence snippets to retrieve (here, we arbitrarily set this to 20). 
By contrast, the LLM approach implicitly and dynamically adjusts this threshold. 
We refer to this baseline as CBERT.

\begin{table}
\small
    \centering
    \footnotesize
    \begin{tabular}{lcccc}
        \hline
\textbf{Diagnosis}&\textbf{Notes}&\multicolumn{2}{c}{\textbf{Evidence}}\\
        &&Flan-&Mistral-\\
        &&T5&Instruct\\
        \hline
        \textit{MIMIC-III}\\
        \hspace{2mm}intracranial hemorrhage*&$95$&$29$&$26$\\
        \hspace{2mm}stroke&$16$&$4$&$2$\\
        \hspace{2mm}small vessel disease&$16$&$8$&$2$\\
        \hspace{2mm}pneumocephalus&$12$&$12$&$11$\\
        \hspace{2mm}sinusitis&$49$&$14$&$3$\\
        \hspace{2mm}\textbf{Total}&\textbf{188}&\textbf{67}&\textbf{44}\\

        \textit{BWH}\\
        \hspace{2mm}small vessel disease&$13$&$8$&$2$\\
        \hspace{2mm}chemoradiation necrosis&$18$&$10$&$20$\\
        \hspace{2mm}demyelination&$21$&$12$&$9$\\
        \hspace{2mm}brain tumor&$21$&$20$&$17$\\
        \hspace{2mm}intracranial hypotension&$20$&$20$&$5$\\
        \hspace{2mm}craniopharyngioma&$20$&$18$&$10$\\
        \hspace{2mm}cerebral infarction&$14$&$14$&$20$\\
        \hspace{2mm}sinusitis&$17$&$15$&$8$\\
        \hspace{2mm}\textbf{Total}&\textbf{144}&\textbf{117}&\textbf{91}\\
        \hline
    \end{tabular}
    \caption{\textbf{Evaluation dataset statistics}. *intracranial hemorrhage is the only diagnosis with more than one patient (it has 4).} 
    \label{tab:dataset}

\end{table}

\begin{figure}
    \centering
    \includegraphics[scale=.35]{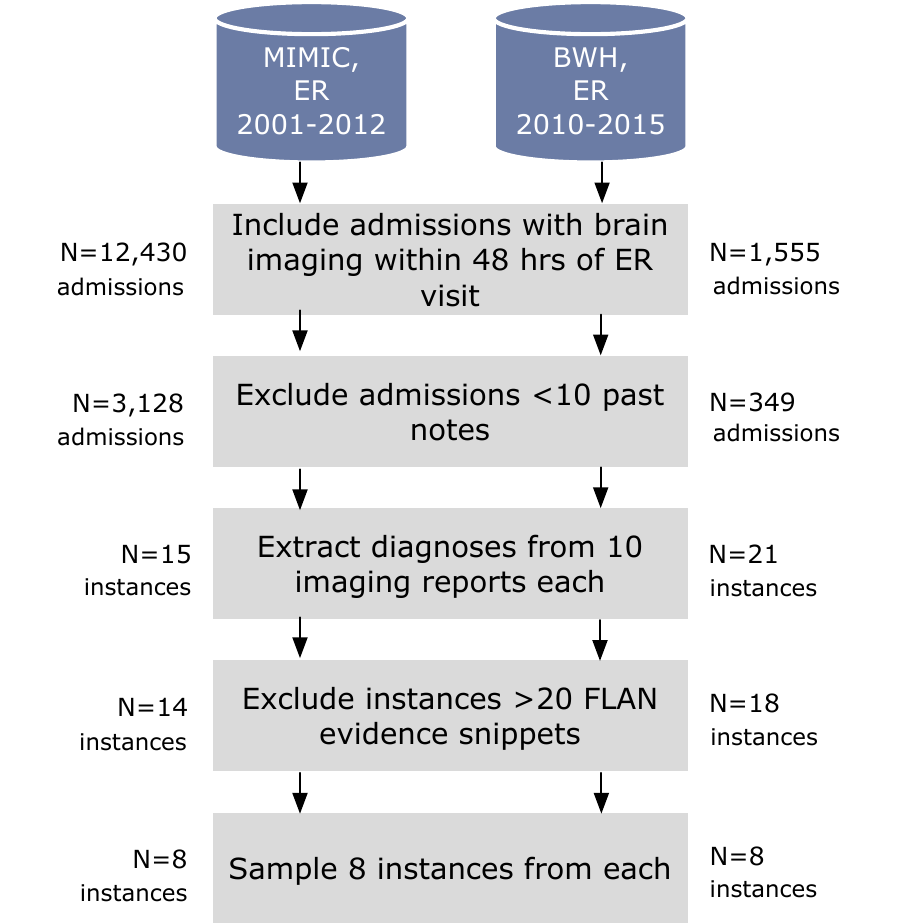}
    \caption{Data sampling flow-chart. An instance is  a unique (patient, diagnosis) combination.} 
    \label{fig:data-sampling}
\end{figure}

\section{Data}
\label{section:data}

For evaluation, we collaborated with radiologists (specializing in neuroimaging) from the Brigham and Women's Hospital in Boston (BWH). Three radiologists with 25,15, and 8 years of experience, respectively, participated in the evaluation. One of them (25 years of experience) had prior experience with LLM projects while the other two did not. 
For experiments, we used a private dataset from 
this hospital and the publicly available MIMIC-III~\citep{johnson2016mimic} dataset, to ensure that our findings are robust and (partially) reproducible.


\vspace{0.35em}
\noindent {\bf BWH} dataset comprises patients admitted to the Emergency Room (ER) of BWH between 2010 and 2015 along with clinical notes including: cardiology, endoscopy, operative, pathology, pulmonary, radiology reports, and discharge summaries. 
We sampled patients who underwent brain imaging within 48 hours of their ER visit. It is typically in the ER that evidence can be beneficial as the initial diagnosis is undetermined in most ER cases. Clinicians attend to several patients in one shift and have to go through often previously unknown patient history to come up with a diagnosis \citep{murray2021medknowts}. Without this constraint, the probability of the diagnosis already being determined in the past would be higher and would trivialize the problem.  
We are interested in scenarios where patients are associated with a large volume of EHR data, so we included patients with $\geq$10 EHR notes.  

\vspace{0.35em}
\noindent {\bf MIMIC-III} is a publicly available database of deidentified EHR from patients admitted to the Intensive Care Unit (ICU) of the Beth Israel Deaconess Medical Center between $2001$ and $2012$. 
It contains both structured data (e.g, demographics, vital sign measurements, lab test results), and unstructured data (e.g.,  nurse and physician notes, ECG and radiology reports and discharge summaries). 
Similar to the BWH dataset, we sampled patients that underwent brain imaging 
 within 48 hours of their ER or Urgent Care visit, whose EHR included $\geq 10$ notes. 


We sampled data for individual patients, but evaluated models with respect to diagnoses. 
For example, if a patient  report mentioned `stroke' and `sinusitis', 
the radiologist evaluated the surfaced evidence for each condition independently. 
To reduce annotation effort, we discarded diagnoses with more than 20 pieces of evidence and finally sampled 8 instances from each source to create our final evaluation dataset.
See Figure \ref{fig:data-sampling} for a schematic of our data sampling procedure. 
Table \ref{tab:dataset} reports statistics about the set of examples used for evaluation.

\begin{table*}[t]
    \centering
    \scalebox{0.95}{
    \footnotesize
    \begin{tabular}{p{1.7cm}p{2cm}p{5.15cm}p{6cm}} 
    \textbf{Evaluation}&\textbf{Diagnosis}&\textbf{Evidence}&\textbf{Explanation}\\
    \hline
    Very Useful&intracranial hemorrhage&Recent fossa surgery and now on anticoagulants&Surgery in the brain inevitably leaves some hemorrhage. Anticoagulants increase the risk of hemorrhage. `Recent surgery' and `anticoagulants' make hemorrhage highly likely.\\
    \hline
    Useful&cerebral infarction&There is calcified thrombus obstructing the origins of the M2 branches&`Thrombus' is diagnostic of infarction, which is very useful information. But `calcified thrombus' implies chronicity, so the thrombus could have been present for a long time and there may not be an acute infarction at this time.\\
    \hline
    Partially Useful&chemoradiation necrosis&\textcolor{red}{The patient is at risk of chemoradiation necrosis due to her history of seizures and brain abscess, which may have caused damage to the brain tissue}. Additionally, her use of concurrent Temodar and involved field radiation during her treatment may have further increased her risk.&History of seizures and brain abscess are not relevant to chemoradiation necrosis. Concurrent  Temodar use and involved field radiation is useful information.\\
    \hline
    Weak Correlation&pneumocephalus&patient was involved in a motorcycle accident&A traumatic head injury is an important risk factor of pneumocephalus. A motorcycle accident increases the likelihood of a head injury.\\
    \hline
    Not Useful&small vessel disease (SVD)&patient is at risk of endocarditis&Not helpful in diagnosing SVD.\\
    \hline
    Hallucination&intracranial hemorrhage&patient has a brain tumor&Not present in the note.
    \end{tabular}}
    \caption{Examples of evidence surfaced by Flan-T5 and Mistral-Instruct for different evaluation categories. Snippet highlighted in \textcolor{red}{red} is irrelevant to the query diagnosis.}
    \label{tab:eval_examples}
\end{table*}

For expert evaluation, one of the collaborating radiologists identified all diagnoses discussed in the \emph{Findings} and \emph{Impressions} sections of the radiology reports of 10 patients from each dataset (excluding MIMIC-III patients from the pilot study).\footnote{While this is a relatively small number of patients, we emphasize that manual evaluation is expensive: Radiologists on our team spent $\sim$9 hours manually assessing outputs.}
Then, for each diagnosis, we retrieved supporting evidence from all patient notes using the zero-shot prompting strategy from Section \ref{section:approach}. 
The three collaborating radiologists then manually assessed each retrieved piece of evidence.

Figure \ref{fig:interface} shows the evaluation interface that our radiologist team-members used to assess model outputs. 
Because the relevance of an evidence snippet inherently depends on the context, 
we ask radiologists to ground their assessments by assuming the following hypothetical setting: ``You are a radiologist reviewing a scan of a patient in the ER. Based on the scan, you are concerned that the patient has the diagnosis stated below. Assess the relevance of the retrieved evidence to support your inference.'' 
For each piece of evidence surfaced by a model, radiologists answered two questions:

\vspace{0.5em}
\noindent \textit{Is the evidence present in the note?} 
    LLMs can hallucinate evidence. 
    Therefore, we first ask radiologists to confirm whether the model generated evidence is in fact supported by the note on the basis of which it was produced. To aid the radiologists in finding the corresponding sentences, we compute ClinicalBERT \citep{alsentzer-etal-2019-publicly} embeddings of sentences in the notes and highlight those 
    with a cosine similarity of $\geq0.9$ with the ClinicalBERT embedding of the generated evidence. 
    This heuristic approach 
    realizes high precision but low recall.
    Therefore, if a highlighted sentence is incongruous with generated evidence, we ask radiologists to read through the entire note to try and manually identify support. 
    
    Note that the (non-generative) retrieval method to which we compare as a baseline is extractive, and so incapable of hallucinating content; we nevertheless ask this question with regards to the baseline for consistency and to ensure blinding. 

    \vspace{0.5em}
    \noindent \textit{Is the evidence relevant?} If generated evidence is supported by the note, we ask radiologists whether it is \emph{relevant} to the query diagnosis. A piece of evidence can contain multiple reasons summarized from across the note. We collect assessments on the following scale (see Table \ref{tab:eval_examples} for examples).
    
    \vspace{0.25em}
    \noindent {\bf Not Useful} \hspace{.1em} None of the evidence is useful; it is irrelevant to the query condition. 
    
    \vspace{0.25em}
    \noindent  {\bf Weak Correlation} \hspace{.1em} Evidence produced has a plausible but weak correlation with the query condition. 
    
   \vspace{0.25em}
    \noindent  {\bf Partially Useful} \hspace{.1em} Out of the multiple risks or signs in the evidence, only some are relevant.
    
    \vspace{0.25em}
    \noindent {\bf Useful} \hspace{.1em} The evidence is relevant and may inform one's diagnostic assessment. 
    
    \vspace{0.25em}
    \noindent  {\bf Very Useful} \hspace{.1em} The evidence is clearly relevant and would likely inform diagnosis.

\section{Results}
\label{section:results}

\begin{figure}
    \centering
    \includegraphics[scale=0.03]{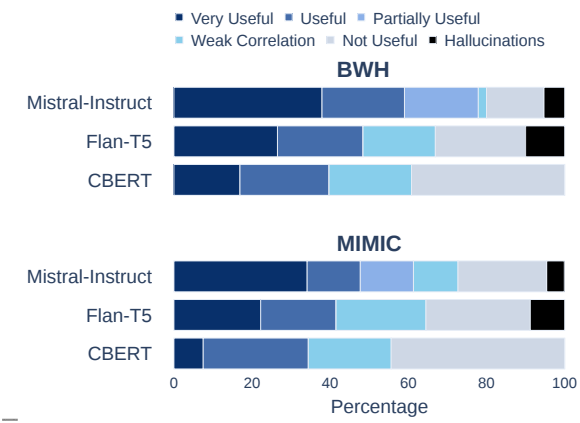}
    \caption{Evidence generated by the LLMs is more often deemed useful than that retrieved by CBERT. 
    But on average, $9.4\%$ and $4.9\%$ of evidence by Flan-T5 and Mistral-Instruct respectively are hallucinated.}
    \label{fig:eval_bar_chart}
\end{figure}

\begin{figure*}[t]
  {%
    \subfigure[Flan-T5]{\label{fig:flan_certainty}%
      \includegraphics[scale=0.375]{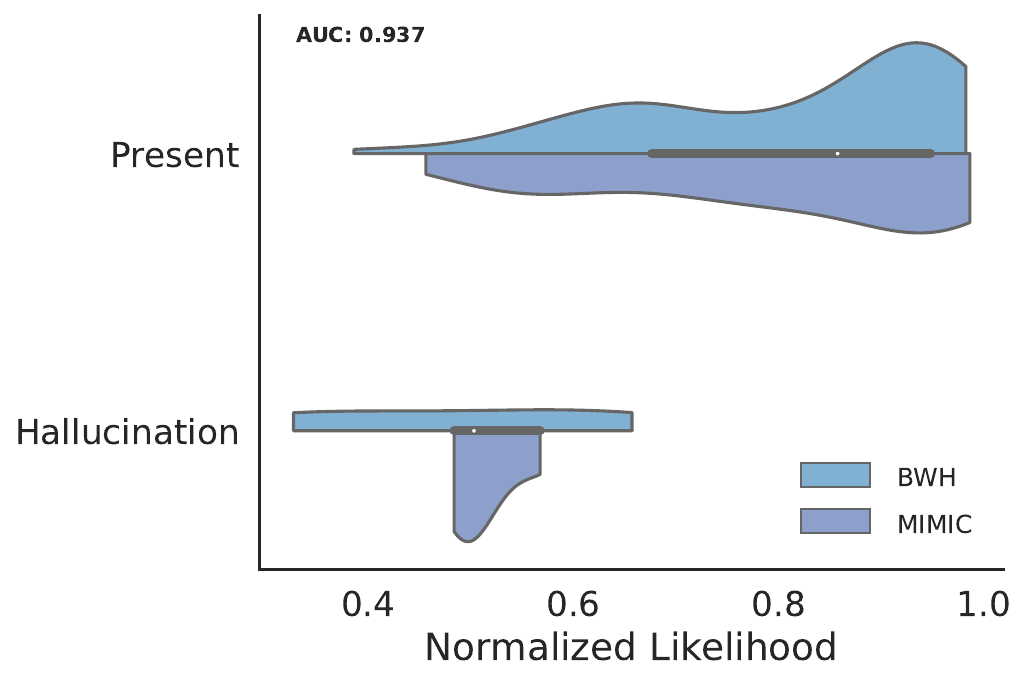}}%
    \qquad
    \subfigure[Mistral-Instruct]{\label{fig:mistral_certainty}%
      \includegraphics[scale=0.375]{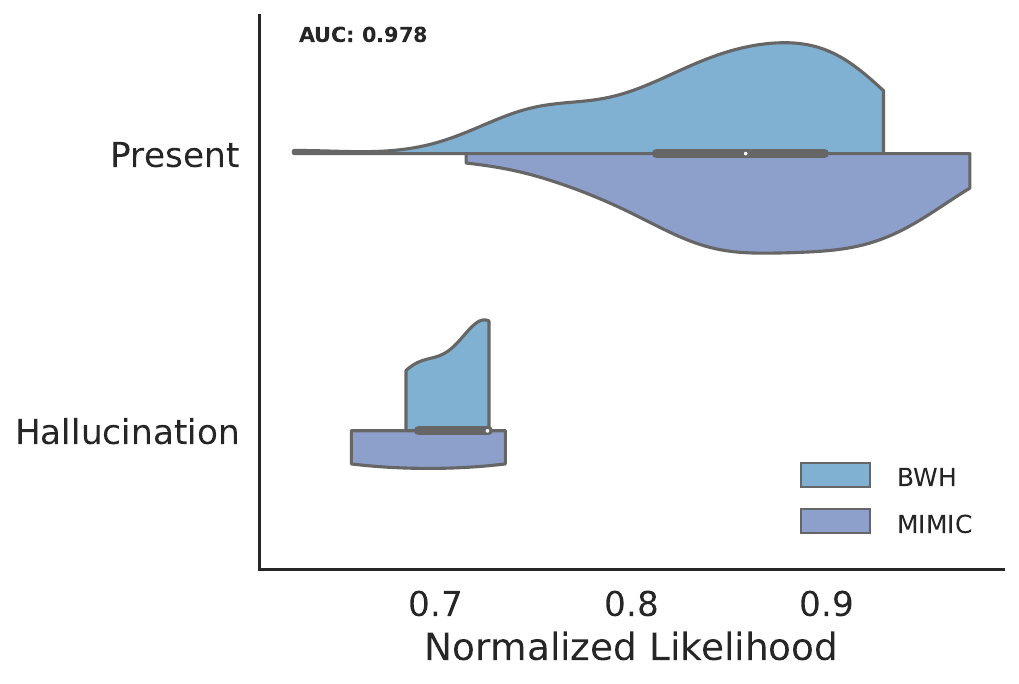}}
  }
  {\caption{Distributions of normalized likelihood, for present and hallucinated evidence. The score provides good discrimination of ``hallucinated'' evidence from present evidence (yielding AUCs of $>$0.9).}\label{fig:lm_certainty}}
\end{figure*} 

To first assess agreement between radiologists, we had all of them annotate evidence surfaced by the LLM for one particular patient, selected at random from the BWH dataset.
For this patient, the model generated 10 pieces of (potentially) relevant evidence for the query \emph{chemoradiation necrosis}. 
On this shared set, the inter-annotator agreement score (average pairwise Cohen's $\kappa$) for relevance assessments between the three radiologists was $0.68$.

Figure \ref{fig:eval_bar_chart} shows our main results. Radiologists found evidence generated by Mistral-Instruct to be the most useful (MIMIC-$47.7\%$, BWH-$59.0\%$), followed by Flan-T5 (MIMIC-$41.5\%$, BWH-$48.4\%$) and then CBERT (MIMIC-$34.4\%$, BWH-$39.7\%$). 
Flan-T5 and CBERT generated more weak correlations than Mistral-Instruct. Both generative models hallucinated evidence. We observed that unlike Mistral-Instruct, Flan-T5 did not summarize multiple reasons from \textit{across} the note as evidence. Hence, none of its evidence was evaluated to be Partially Useful. Since CBERT is extractive, there is no clear indication of which condition is to be evaluated as evidence. For this reason, the evidence from CBERT was evaluated overall and Partially Useful was not used.
The assessment of generated evidence implicitly measures precision. We also estimate model recall in \S \ref{app:recall}.

\subsection{Hallucinations} 
\label{sec:hallucinations}

Concerningly, some model hallucinations flagged by radiologists include plausible risk factors. 
A few illustrative examples:

\vspace{0.25em}
\noindent {\bf Example 1} For a patient with demyelination as the query diagnosis, Flan-T5 hallucinated the evidence `axonal degeneration'. Demyelination is commonly viewed as the primary factor responsible for the deterioration of axons within multiple sclerosis lesions. The model also hallucinated signs of demyelination as evidence (`numbness and tingling in the arms and legs'). There was no evidence 
indicating axonal degeneration or the symptoms. 
    
\vspace{0.25em}
\noindent {\bf Example 2} For a patient with chemoradiation necrosis as the query diagnosis, Mistral-Instruct hallucinated that `the patient had a history of chemoradiation necrosis'. A history of chemoradiation necrosis would be very relevant to its diagnosis, but there was no such history in the EHR.

\vspace{0.25em}
\noindent In other instances, the model hallucinated vague evidence, e.g., `The patient is taking a lot of medications that can cause small vessel disease' for small vessel disease as the query diagnosis (
a radiologist went through the note and was unable to find mention of any such medication). 


\paragraph{How certain is the model about such hallucinations?}
\label{sec:certainty}
We evaluate the degree to which model uncertainty---normalized output likelihoods under the LM---suggests `hallucinated' content (Figure \ref{fig:lm_certainty}).  
Both models considered yield confidence scores that are highly indicative of hallucinations.
This is promising, as it suggests we can simply abstain from providing outputs in such cases. 

\subsection{Weakly correlating evidence} 
\label{section:weak}
A factor complicating evaluation is that LLMs often yield evidence which has plausible but weak correlation with a query condition.  
One could argue that the model was `correct' in retrieving such evidence from an epidemiology perspective, but incorrect (or at least not useful) from an individual patient, clinical perspective. 
In other words, evidence may be so weakly correlated with a condition that it is of small value, even if technically `correct'. 
See Tables \ref{tab:eval_examples} and \ref{tab:weak_corr_examples} for examples.

\begin{table*}
\centering
\footnotesize
\begin{tabular}{p{8cm}|p{8cm}}
\multicolumn{1}{c|}{\textbf{FLAN-T5/Mistral-Instruct}}&\multicolumn{1}{c}{\textbf{CBERT}}\\
\hline
CAD (s/p stents x 2, $>2$ MIs, on Coumadin INR=1.9) hx of $>3$ TIAs in past 2.5 yrs multiple AAAs (largest last measured at 5.5 cm, surg intervention held 2/2 cardiac status&IMMUNIZATIONS: INFLUENZA VACCINE (INACTIVATED) IM Given [DATE] ALLERGY: AMOXICILLIN ADMIT DIAGNOSIS: Stroke PRINCIPAL DISCHARGE DIAGNOSIS ;Responsible After Study for Causing Admission) same OTHER DIAGNOSIS;Conditions,Infections,Complications,affecting Treatment/Stay CAD (s/p stents x 2, $>2$ MIs, on Coumadin INR=1.9) hx of $>3$ TIAs in past 2.5 yrs multiple AAAs (largest last measured at 5.5 cm, surg intervention held 2/2 cardiac status OPERATIONS AND PROCEDURES: None.OTHER TREATMENTS/PROCEDURES (NOT IN O.R.)\\
\hline
The patient has a TBI&A/P- S/P REPAIR [**Doctor Last Name **] \& LL ORTHOPEDIC INJURIES STABLE TBI W/CLOSE MONITORING FOR CHANGES STABLE LIVER LAC AT PRESENT SUCCESSFULL WEAN/EXTUBATION  POST-OP PAIN  CONT TO MONITOR PER ORDERS- Q2/HR NEURO \& PERIPHERAL VASCULAR CHECKS...?\\
\hline
\end{tabular}
\caption{Examples of evidence when generative models are more concise than CBERT, highlighting the benefits of abstractive summarization.}
\label{tab:conciseness_examples}
\end{table*}

\subsection{Qualitative Evaluation}
We summarize the comments offered by radiologists during evaluation. Radiologists found outputs of Mistral-Instruct and FLAN-T5 to be more precise and concise compared to CBERT. Abstractive evidence was considered better than the extractive snippets from CBERT, which often chunked useful evidence with neighboring irrelevant sentences (notes are usually poorly formatted, making sentence-parsing difficult). See Table \ref{tab:conciseness_examples} for examples. CBERT was preferred in three cases, when both Mistral-Instruct and FLAN-T5 had poor precision or recall. For instance, both models had a precision of $\sim50\%$ for pneumocephalus. Interestingly, our radiologist preferred CBERT for the case of demyelination because it helped confirm that the patient did $not$ have demyelination, but in fact had a glioma (tumor). Demyelinating lesions and glioma present similar imaging characteristics and can be difficult to diagnose based on conventional MR imaging \citep{toh2012differentiation}. A brain biopsy is often conducted to differentiate between the two. All the evidence evaluated as (very) useful were snippets from the pathology report discussing the tests and related results that indicated that demyelination was less likely and that the findings were most consistent with glioma.

\begin{table}[h]
    \centering
    \small
    \begin{tabular}{lcccc}
    \hline
    Model&\multicolumn{2}{c}{MIMIC}&\multicolumn{2}{c}{BWH}\\
    \hline
    \multicolumn{5}{c}{\textbf{2. Verify presence of each risk factor/sign}}\\
    &H&P&H&P\\
    Flan-T5&$75.0(4)$&$90.0$&$83.3(6)$&$86.1$\\
    Mistral-Inst.&$100.0(3)$&$88.2$&$60.0(5)$&$95.1$\\
    & & & & \\
    \multicolumn{5}{c}{\textbf{3. Check validity of present risk factors/signs}}\\
    &F1&PCC&F1&PCC\\
    Flan-T5&$75.6$&$79.2$&$74.2$&$37.8$\\
    Mistral-Inst.&$81.4$&$92.0$&$77.5$&$34.2$\\
    CBERT&$55.0$&$41.1$&$63.9$&$68.1$\\
    \hline
    \end{tabular}
    \caption{Evaluating automatic evaluation. We first compute the accuracy for hallucinated (H) and present (P) evidence (Step 2 in Figure \ref{fig:auto_eval_flow}). We then compute micro-F1 and PCC for present evidence (Step 3 in Figure \ref{fig:auto_eval_flow}).}
    \label{tab:auto-eval}
\end{table}

\begin{figure}[h]
    \centering
    \includegraphics[scale=0.35]{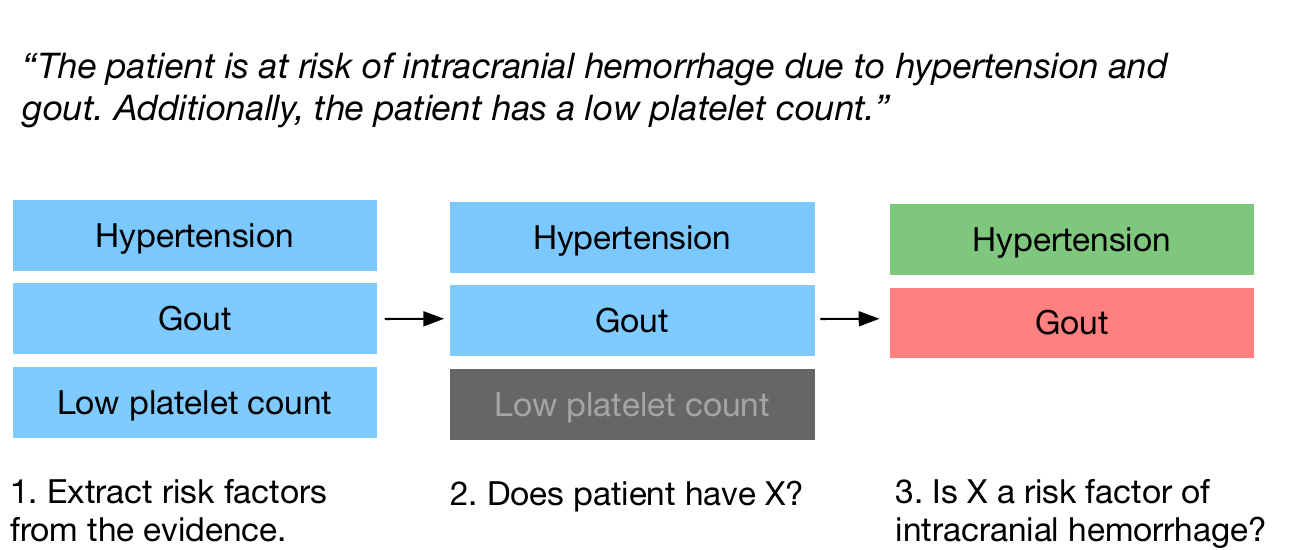}
    \caption{Automatic LLM-based evaluation of retrieved evidence. The evaluator LLM: (1) extracts risk factors from the evidence; (2) verifies the presence of each in the note; and (3) validates each present risk factor. The same approach is adopted for evaluating signs of the query diagnosis.}
    \label{fig:auto_eval_flow}
\end{figure}
\section{Automatic Evaluation}
\label{sec:auto-eval}

Manually evaluating evidence requires a considerable amount of scarce (expensive) expert time, meaning it does not scale. 
This limited our evaluation above to a small set of patients.  
To expand our evaluation we now also consider the use of LLMs as evaluators. 
Prior work has established that LLM-based evaluation can provide meaningful signal in general \citep{chiang2023can, min2023factscore, chang2023booookscore, kim2023prometheus}, but there has been limited work investigating such evaluation in  healthcare; 
it is important to assess automatic evaluation in this domain due to the high cost of manual annotation.  

In this section, we first verify the degree to which LLM-based automatic evaluations correlate with manual (expert) assessments (\S \ref{section:eval-eval}).
Finding evidence of meaningful (if noisy) correlation, we then use this automated approach to increase the scale of our evaluation (\S \ref{section:scaling-eval}).

Figure \ref{fig:auto_eval_flow} provides an overview of our approach. 
Given a piece of evidence generated by an LLM to evaluate, we use an evaluator LLM to: (1) Extract the risk factors it contains; (2) Verify the presence of each risk factor in the note; (3) Check if each present risk factor is a valid risk factor of the query diagnosis.
We execute these steps sequentially by one-shot prompting the evaluator LLM for (1) and zero-shot prompting it for (2) and (3). We provide more details in \S \ref{app:auto_eval}. 
Note that steps (2) and (3) are performed separately for each extracted risk factor. 
Recall that in addition to risk factors, we prompt for signs of diagnosis; we follow the same approach to evaluate these. 

\subsection{Evaluating automatic evaluation}
\label{section:eval-eval}

We first validate this automated (LLM-based) evaluation approach for our task by comparing it to the expert evaluations described in \S \ref{section:data}. 
Given its superior performance according to expert evaluations, we use Mistral-Instruct as the LLM evaluator. 
We compute micro-F1 and Pearson's Correlation Coefficient (PCC), using expert evaluations on the set of instances manually annotated as the ground truth. 
Micro-F1 measures how well the LLM evaluates each extracted risk or sign individually (irrespective of which instance 
these are associated with). 
PCC is computed at the \emph{instance-level} by calculating the average relevance over extracted risks and signs from all pieces of evidence; this is therefore an aggregate measure of how well the LLM evaluates an instance. 

Because automatic evaluation yields binary predictions (whether a risk factor/sign is relevant to the diagnosis or not), we map expert \emph{relevance} scale to binary labels: Not Useful $\rightarrow0$ and \{Weak Correlation, Useful, Very Useful\} $\rightarrow1$. 
For \emph{evidence}, we assign $1$ to pieces marked as (Very) Useful or Weak Correlations, and $0$ to the rest. 
As discussed in \S \ref{section:weak}, Weak Correlations fall into a grey area. 
Therefore, we also perform a \textit{strict} evaluation where Weak Correlations $\rightarrow0$. 
We report results in Table \ref{tab:auto-eval}, and offer the following observations. 

\vspace{0.3em}
\noindent \textbf{Hallucinations can be automatically detected.} As seen in Table \ref{tab:auto-eval} (top), prompting to confirm whether a patient has a condition based on the note permits discrimination of ``hallucinated'' and actually present conditions. 

\vspace{0.3em}
\noindent \textbf{Micro-F1 scores are high for generative evidence.}  
The evaluator LLM is able to extract and validate risk factors and signs of diagnoses in a way that agrees reasonably well with human experts. 
The micro-F1 scores are high for both Flan-T5 and Mistral-Instruct across the datasets. 

\vspace{0.3em}
\noindent \textbf{Micro-F1 scores are relatively low for the baseline retrieval approach.} 
CBERT fares comparatively poorly here. Prompting for risk factors and signs from extractive evidence is difficult because these are not as explicitly stated (as opposed to generative outputs of the format `The patient is at risk of $X$ because of $Y$') and are buried in irrelevant information. (This issue was observed during expert evaluation as well.) The result is noisy outputs (e.g., `intubation', `worsening respiratory status', `age') that generate false positives for valid risk factors and signs. 
This highlights the relative advantage of LLMs for flexible evidence retrieval. 

\vspace{0.25em}
\noindent \textbf{PCC varies from moderate to high.} While PCC is high for both Flan-T5 and Mistral-Instruct for MIMIC, the correlation is moderate for BWH. This is apparently due to poor evaluative performance for one diagnosis (chemoradiation necrosis for Flan-T5 and intracranial hypotension for Mistral-Instruct). In both cases, a unique risk factor was incorrectly validated by the evaluator LLM. But multiple occurrences of the risk factor across notes, resulting in repeated retrieval as evidence, significantly brought down PCC. 
Removing the diagnoses out increases PCC to $82.3$ and $51.3$ for Flan-T5 and Mistral-Instruct, respectively.

\vspace{0.35em}
\noindent \textbf{Correlation drops significantly in strict evaluation.} 
Table \ref{tab:strict-eval} shows the \textit{change} in micro-F1 and PCC when strict evaluation is performed (compared to when Weak Correlations $\rightarrow1$, shown in Table \ref{tab:auto-eval}). 
With the exception of PCC for CBERT (MIMIC), there is a drop in micro-F1 and PCC across all model-dataset combinations when Weak Correlations $\rightarrow0$. This owes to the inherent complexity of evaluating clinical evidence (automatically or otherwise). 
What constitutes `Useful' evidence for supporting diagnosis is, to a degree, inherently subjective.


\begin{table}[h]
    \centering
    \small
    \begin{tabular}{lcc|cc}
    Model&\multicolumn{2}{c}{MIMIC}&\multicolumn{2}{c}{BWH}\\
    &$\Delta$ F1&$\Delta$ PCC&$\Delta$ F1&$\Delta$ PCC\\
    \hline
    Flan-T5&$9.9$\textcolor{red}{$\downarrow$}&$9.8$\textcolor{red}{$\downarrow$}&$6.3$\textcolor{red}{$\downarrow$}&$9.7$\textcolor{red}{$\downarrow$}\\
    Mistral-Instruct&$15.3$\textcolor{red}{$\downarrow$}&$14.8$\textcolor{red}{$\downarrow$}&$1.9$\textcolor{red}{$\downarrow$}&$13.5$\textcolor{red}{$\downarrow$}\\
    CBERT&$14.1$\textcolor{red}{$\downarrow$}&$18.7$\textcolor{limegreen}{$\uparrow$}&$13.5$\textcolor{red}{$\downarrow$}&$13.7$\textcolor{red}{$\downarrow$}\\
    \hline
    \end{tabular}
    \caption{Evaluating \textbf{strict} automatic evaluation metrics. The figures here indicate the \textit{change} in micro-F1 and PCC compared to when Weak Correlations $\rightarrow1$ (shown in Table \ref{tab:auto-eval}). Correlation with expert evaluation drops when Weak Correlations $\rightarrow0$.}
    \label{tab:strict-eval}
\end{table}

\noindent Overall, automatic evaluation using an LLM has a meaningful correlation (micro-F1) with expert evaluation when measured at risk factor (sign)-level. 
At the instance-level, the correlation (PCC) is moderate (BWH) to high (MIMIC). 
The variance may owe to the small number of instances evaluated.

\subsection{Scaling our Evaluation}
\label{section:scaling-eval}

Having verified that automatic evaluation provides an imperfect but meaningful assessment of outputs, we now scale our evaluation using this approach. 
Specifically, we complement our manual analysis with an automatic evaluation of the three models at a larger scale.
We evaluate $100$ and $50$ instances (patient-diagnosis combinations) for MIMIC and BWH respectively. 
As discussed in \S \ref{section:data}, a collaborating radiologist identified the query diagnoses in the radiology reports during manual evaluation. 
For this automatic evaluation, we follow prior work \cite{tang2023less}, and consider conditions following \textit{likely indicators} (such as `concerning for', `diagnosis include'. Details in \S \ref{app:likely_indicators}) as diagnoses. 

Table \ref{tab:scaled-up-eval} shows results of the scaled up evaluation (see Table \ref{tab:scaled-up-eval-stats} for data statistics). 
Both Flan-T5 and Mistral-Instruct significantly outperform CBERT, consistent with the findings from our manual evaluation. 
Mistral-Instruct appears to generate more useful evidence compared to Flan-T5 (again consistent with the manual evaluation). 
Both models have comparable rates of hallucination for MIMIC but Flan-T5 has a higher rate for BWH.

\begin{table}
    \centering
    \small
    \begin{tabular}{lccc}
    Model&Useful&Not Useful&Hallucinations\\
    \hline
    Flan-T5\\
    \hspace{0.8em}MIMIC&$48.5$&$42.1$&$9.4$\\
    \hspace{0.8em}BWH&$47.0$&$38.4$&$14.6$\\
    Mistral-Instruct\\
    \hspace{0.8em}MIMIC&$55.0$&$35.9$&$9.1$\\
    \hspace{0.8em}BWH&$59.8$&$32.0$&$8.2$\\
    CBERT\\
    \hspace{0.8em}MIMIC&$29.7$&$70.3$&-\\
    \hspace{0.8em}BWH&$28.7$&$71.3$&-\\
    \hline
    \end{tabular}
    \caption{Results of large-scale evaluation performed by using Mistral-Instruct as an evaluator. LLMs outperform the retrieval baseline. Mistral-Instruct generates more useful evidence compared to Flan-T5.}
    \label{tab:scaled-up-eval}
\end{table}



\section{Related Work}  
\paragraph{NLP for EHR.}
Navigating EHRs is cumbersome, motivating efforts in summarization of and information extraction from EHR \citep{pivovarov2015automated}. 
For example, in recent related work, \cite{jiang2023conceptualizing} created a proactive note retrieval system based on the current clinical context to aid note-writing. 
\cite{adams2021s} considered ``hospital-course summarization'', condensing the notes of a patient visit into a paragraph. Other work \cite{liang2019novel} has sought to produce disease-specific summaries from notes. 

\paragraph{LLMs for healthcare.} There has been a flurry of work on the capabilities of LLMs for healthcare \emph{generally}, i.e., in terms of ability to answer general questions and take medical exams, e.g., \cite{singhal2023towards,pmlr-v209-eric23a,nori2023can,yang2022large}. Our work, however, is focused on a grounded, specific task. 

\paragraph{NLP in Radiology.}
Previous works regarding NLP in radiology primarily focus on processing radiology reports. Some work has sought to automatically generate the Impression section of reports \citep{van2023radadapt, zhang2019optimizing, sotudeh2020attend}.
Other efforts have focused on extracting specific observations 
\citep{smit2020chexbert, jaiswal2021radbert}, and modeling disease progression \citep{di2021diagnostic,khanna2023radgraph2}.

\paragraph{NLP to aid diagnosis.} The prior work most relevant to this effort concern aiding radiologists in diagnosis.  
McInerney \emph{et al.} (\citeyear{mcinerney2020query}) propose a distantly supervised model (trained to predict ICD codes) to perform extractive summarization conditioned on a diagnoses; our work addresses this problem with LLMs, \emph{zero-shot}. 
\citealt{tang2023less} address diagnostic uncertainty by suggesting less likely diagnosis to radiologists, learnt by differentiating between likely and less likely diagnoses via contrastive learning.


\section{Discussion and Limitations}
\label{sec:conclusions} 

We proposed an approach for using LLMs to retrieve and summarize evidence from patient records which might be relevant to a particular 
diagnosis of interest, with the aim of aiding radiologists performing imaging diagnosis. 
Expert evaluations of model outputs performed by radiologists show that this is a promising approach, as compared to pre-LLM techniques. 
We also established that automated (LLM-based) evaluation is feasible, and confirmed our findings using this approach. 

There are important {\bf limitations} to the approach and to our evaluation.
We found that LLMs are prone to hallucinating (plausible) evidence, potentially hindering their utility for the envisioned use. 
However, our results also indicate that model confidence might allow one to pro-actively identify hallucinations, and abstain from providing (generative) summaries in such cases; extending this is an interesting direction for future work. 

Our evaluation was limited in a few key ways. 
We enlisted radiologists to perform in-depth evaluation of a small number of instances, because evaluation is time consuming: We emphasize that this exercise required substantial allocation ($\sim$16 hours) of scarce expert time. 
We attempted to mitigate this via LLM-based automatic evaluation, performed at larger scale. 
However, our assessment of this strategy also relied on this relatively small annotated set and so may not generalize. 
Another limitation here is that we considered only two LLMs (specifically, FLAN-T5 and Mistral-Instruct): 
Other LLMs might, naturally, perform better or worse. In addition, we did not investigate the fairness implications of our work. However, the small size of our expert-annotated sample and the inherently small samples of underrepresented groups limits our ability to meaningfully assess this. We leave the detailed analysis needed to determine if there are significant differences to future work. Finally, we did not extensively iterate on the prompts used, and this too could substantially affect results.

\section{Acknowledgment}
\label{sec:acknowledgment}
We acknowledge partial funding for this work by National Library of Medicine of the National Institutes of Health (NIH) under award numbers R01LM013772 and R01LM013891.  The content is solely the responsibility of the authors and does not necessarily represent the official views of the NIH.

\bibliography{jmlr-sample}

\begin{thebibliography}{35}
\providecommand{\natexlab}[1]{#1}
\providecommand{\url}[1]{\texttt{#1}}
\expandafter\ifx\csname urlstyle\endcsname\relax
  \providecommand{\doi}[1]{doi: #1}\else
  \providecommand{\doi}{doi: \begingroup \urlstyle{rm}\Url}\fi

\bibitem[Adams et~al.(2021)Adams, Alsentzer, Ketenci, Zucker, and Elhadad]{adams2021s}
Griffin Adams, Emily Alsentzer, Mert Ketenci, Jason Zucker, and No{\'e}mie Elhadad.
\newblock What’s in a summary? laying the groundwork for advances in hospital-course summarization.
\newblock In \emph{Proceedings of the conference. Association for Computational Linguistics. North American Chapter. Meeting}, volume 2021, page 4794. NIH Public Access, 2021.

\bibitem[Agrawal et~al.(2022)Agrawal, Hegselmann, Lang, Kim, and Sontag]{agrawal2022large}
Monica Agrawal, Stefan Hegselmann, Hunter Lang, Yoon Kim, and David Sontag.
\newblock Large language models are zero-shot clinical information extractors.
\newblock \emph{arXiv preprint arXiv:2205.12689}, 2022.

\bibitem[Alsentzer et~al.(2019)Alsentzer, Murphy, Boag, Weng, Jindi, Naumann, and McDermott]{alsentzer-etal-2019-publicly}
Emily Alsentzer, John Murphy, William Boag, Wei-Hung Weng, Di~Jindi, Tristan Naumann, and Matthew McDermott.
\newblock Publicly available clinical {BERT} embeddings.
\newblock In \emph{Proceedings of the 2nd Clinical Natural Language Processing Workshop}, pages 72--78, Minneapolis, Minnesota, USA, June 2019. Association for Computational Linguistics.
\newblock \doi{10.18653/v1/W19-1909}.
\newblock URL \url{https://aclanthology.org/W19-1909}.

\bibitem[Azamfirei et~al.(2023)Azamfirei, Kudchadkar, and Fackler]{azamfirei2023large}
Razvan Azamfirei, Sapna~R Kudchadkar, and James Fackler.
\newblock Large language models and the perils of their hallucinations.
\newblock \emph{Critical Care}, 27\penalty0 (1):\penalty0 1--2, 2023.

\bibitem[Chang et~al.(2023)Chang, Lo, Goyal, and Iyyer]{chang2023booookscore}
Yapei Chang, Kyle Lo, Tanya Goyal, and Mohit Iyyer.
\newblock Booookscore: A systematic exploration of book-length summarization in the era of llms.
\newblock \emph{arXiv preprint arXiv:2310.00785}, 2023.

\bibitem[Chiang and Lee(2023)]{chiang2023can}
Cheng-Han Chiang and Hung-yi Lee.
\newblock Can large language models be an alternative to human evaluations?
\newblock \emph{arXiv preprint arXiv:2305.01937}, 2023.

\bibitem[Chung et~al.(2022)Chung, Hou, Longpre, Zoph, Tay, Fedus, Li, Wang, Dehghani, Brahma, et~al.]{chung2022scaling}
Hyung~Won Chung, Le~Hou, Shayne Longpre, Barret Zoph, Yi~Tay, William Fedus, Eric Li, Xuezhi Wang, Mostafa Dehghani, Siddhartha Brahma, et~al.
\newblock Scaling instruction-finetuned language models.
\newblock \emph{arXiv preprint arXiv:2210.11416}, 2022.

\bibitem[Di~Noto et~al.(2021)Di~Noto, Atat, Teiga, Hegi, Hottinger, Cuadra, Hagmann, and Richiardi]{di2021diagnostic}
Tommaso Di~Noto, Chirine Atat, Eduardo~Gamito Teiga, Monika Hegi, Andreas Hottinger, Meritxell~Bach Cuadra, Patric Hagmann, and Jonas Richiardi.
\newblock Diagnostic surveillance of high-grade gliomas: towards automated change detection using radiology report classification.
\newblock In \emph{Joint European Conference on Machine Learning and Knowledge Discovery in Databases}, pages 423--436. Springer, 2021.

\bibitem[Honnibal and Montani(2017)]{spacy2}
Matthew Honnibal and Ines Montani.
\newblock {spaCy 2}: Natural language understanding with {B}loom embeddings, convolutional neural networks and incremental parsing.
\newblock To appear, 2017.

\bibitem[Jaiswal et~al.(2021)Jaiswal, Tang, Ghosh, Rousseau, Peng, and Ding]{jaiswal2021radbert}
Ajay Jaiswal, Liyan Tang, Meheli Ghosh, Justin~F Rousseau, Yifan Peng, and Ying Ding.
\newblock Radbert-cl: Factually-aware contrastive learning for radiology report classification.
\newblock In \emph{Machine Learning for Health}, pages 196--208. PMLR, 2021.

\bibitem[Jiang et~al.(2023{\natexlab{a}})Jiang, Sablayrolles, Mensch, Bamford, Chaplot, de~las Casas, Bressand, Lengyel, Lample, Saulnier, Lavaud, Lachaux, Stock, Scao, Lavril, Wang, Lacroix, and Sayed]{jiang2023mistral}
Albert~Q. Jiang, Alexandre Sablayrolles, Arthur Mensch, Chris Bamford, Devendra~Singh Chaplot, Diego de~las Casas, Florian Bressand, Gianna Lengyel, Guillaume Lample, Lucile Saulnier, Lélio~Renard Lavaud, Marie-Anne Lachaux, Pierre Stock, Teven~Le Scao, Thibaut Lavril, Thomas Wang, Timothée Lacroix, and William~El Sayed.
\newblock Mistral 7b, 2023{\natexlab{a}}.

\bibitem[Jiang et~al.(2023{\natexlab{b}})Jiang, Shen, Agrawal, Lam, Kurtzman, Horng, Karger, and Sontag]{jiang2023conceptualizing}
Sharon Jiang, Shannon Shen, Monica Agrawal, Barbara Lam, Nicholas Kurtzman, Steven Horng, David Karger, and David Sontag.
\newblock Conceptualizing machine learning for dynamic information retrieval of electronic health record notes.
\newblock \emph{arXiv preprint arXiv:2308.08494}, 2023{\natexlab{b}}.

\bibitem[Johnson et~al.(2016{\natexlab{a}})Johnson, Pollard, and Mark]{johnson2016physionet}
Alistair E.~W. Johnson, Tom~J. Pollard, and Roger~G. Mark.
\newblock {MIMIC-III} clinical database (version 1.4), 2016{\natexlab{a}}.

\bibitem[Johnson et~al.(2016{\natexlab{b}})Johnson, Pollard, Shen, Lehman, Feng, Ghassemi, Moody, Szolovits, Anthony~Celi, and Mark]{johnson2016mimic}
Alistair~EW Johnson, Tom~J Pollard, Lu~Shen, Li-wei~H Lehman, Mengling Feng, Mohammad Ghassemi, Benjamin Moody, Peter Szolovits, Leo Anthony~Celi, and Roger~G Mark.
\newblock Mimic-iii, a freely accessible critical care database.
\newblock \emph{Scientific data}, 3\penalty0 (1):\penalty0 1--9, 2016{\natexlab{b}}.

\bibitem[Khanna et~al.(2023)Khanna, Dejl, Yoon, Truong, Duong, Saenz, and Rajpurkar]{khanna2023radgraph2}
Sameer Khanna, Adam Dejl, Kibo Yoon, Quoc~Hung Truong, Hanh Duong, Agustina Saenz, and Pranav Rajpurkar.
\newblock Radgraph2: Modeling disease progression in radiology reports via hierarchical information extraction.
\newblock \emph{arXiv preprint arXiv:2308.05046}, 2023.

\bibitem[Kim et~al.(2023)Kim, Shin, Cho, Jang, Longpre, Lee, Yun, Shin, Kim, Thorne, et~al.]{kim2023prometheus}
Seungone Kim, Jamin Shin, Yejin Cho, Joel Jang, Shayne Longpre, Hwaran Lee, Sangdoo Yun, Seongjin Shin, Sungdong Kim, James Thorne, et~al.
\newblock Prometheus: Inducing fine-grained evaluation capability in language models.
\newblock \emph{arXiv preprint arXiv:2310.08491}, 2023.

\bibitem[Lehman et~al.(2023)Lehman, Hernandez, Mahajan, Wulff, Smith, Ziegler, Nadler, Szolovits, Johnson, and Alsentzer]{pmlr-v209-eric23a}
Eric Lehman, Evan Hernandez, Diwakar Mahajan, Jonas Wulff, Micah~J Smith, Zachary Ziegler, Daniel Nadler, Peter Szolovits, Alistair Johnson, and Emily Alsentzer.
\newblock Do we still need clinical language models?
\newblock In Bobak~J. Mortazavi, Tasmie Sarker, Andrew Beam, and Joyce~C. Ho, editors, \emph{Proceedings of the Conference on Health, Inference, and Learning}, volume 209 of \emph{Proceedings of Machine Learning Research}, pages 578--597. PMLR, 22 Jun--24 Jun 2023.
\newblock URL \url{https://proceedings.mlr.press/v209/eric23a.html}.

\bibitem[Liang et~al.(2019)Liang, Tsou, and Poddar]{liang2019novel}
Jennifer Liang, Ching-Huei Tsou, and Ananya Poddar.
\newblock A novel system for extractive clinical note summarization using ehr data.
\newblock In \emph{Proceedings of the Clinical Natural Language Processing Workshop}, pages 46--54, 2019.

\bibitem[McInerney et~al.(2020)McInerney, Dabiri, Touret, Young, Meent, and Wallace]{mcinerney2020query}
Denis~Jered McInerney, Borna Dabiri, Anne-Sophie Touret, Geoffrey Young, Jan-Willem Meent, and Byron~C Wallace.
\newblock Query-focused ehr summarization to aid imaging diagnosis.
\newblock In \emph{Machine Learning for Healthcare Conference}, pages 632--659. PMLR, 2020.

\bibitem[McInerney et~al.(2023)McInerney, Young, van~de Meent, and Wallace]{jered-emnlp-23}
Denis~Jered McInerney, Geoffrey Young, Jan-Willem van~de Meent, and Byron~C. Wallace.
\newblock {CHiLL: Zero-shot Custom Interpretable Feature Extraction from Clinical Notes with Large Language Models}.
\newblock In \emph{Proceeding of Findings of the Conference on Empirical Methods for Natural Language Processing (EMNLP)}, 2023.

\bibitem[Min et~al.(2023)Min, Krishna, Lyu, Lewis, Yih, Koh, Iyyer, Zettlemoyer, and Hajishirzi]{min2023factscore}
Sewon Min, Kalpesh Krishna, Xinxi Lyu, Mike Lewis, Wen-tau Yih, Pang~Wei Koh, Mohit Iyyer, Luke Zettlemoyer, and Hannaneh Hajishirzi.
\newblock Factscore: Fine-grained atomic evaluation of factual precision in long form text generation.
\newblock \emph{arXiv preprint arXiv:2305.14251}, 2023.

\bibitem[Murray et~al.(2021)Murray, Gopinath, Agrawal, Horng, Sontag, and Karger]{murray2021medknowts}
Luke Murray, Divya Gopinath, Monica Agrawal, Steven Horng, David Sontag, and David~R Karger.
\newblock Medknowts: unified documentation and information retrieval for electronic health records.
\newblock In \emph{The 34th Annual ACM Symposium on User Interface Software and Technology}, pages 1169--1183, 2021.

\bibitem[Nori et~al.(2023)Nori, Lee, Zhang, Carignan, Edgar, Fusi, King, Larson, Li, Liu, Luo, McKinney, Ness, Poon, Qin, Usuyama, White, and Horvitz]{nori2023can}
Harsha Nori, Yin~Tat Lee, Sheng Zhang, Dean Carignan, Richard Edgar, Nicolo Fusi, Nicholas King, Jonathan Larson, Yuanzhi Li, Weishung Liu, Renqian Luo, Scott~Mayer McKinney, Robert~Osazuwa Ness, Hoifung Poon, Tao Qin, Naoto Usuyama, Chris White, and Eric Horvitz.
\newblock Can generalist foundation models outcompete special-purpose tuning? case study in medicine.
\newblock November 2023.

\bibitem[Pivovarov and Elhadad(2015)]{pivovarov2015automated}
Rimma Pivovarov and No{\'e}mie Elhadad.
\newblock Automated methods for the summarization of electronic health records.
\newblock \emph{Journal of the American Medical Informatics Association}, 22\penalty0 (5):\penalty0 938--947, 2015.

\bibitem[Singhal et~al.(2023)Singhal, Tu, Gottweis, Sayres, Wulczyn, Hou, Clark, Pfohl, Cole-Lewis, Neal, et~al.]{singhal2023towards}
Karan Singhal, Tao Tu, Juraj Gottweis, Rory Sayres, Ellery Wulczyn, Le~Hou, Kevin Clark, Stephen Pfohl, Heather Cole-Lewis, Darlene Neal, et~al.
\newblock Towards expert-level medical question answering with large language models.
\newblock \emph{arXiv preprint arXiv:2305.09617}, 2023.

\bibitem[Smit et~al.(2020)Smit, Jain, Rajpurkar, Pareek, Ng, and Lungren]{smit2020chexbert}
Akshay Smit, Saahil Jain, Pranav Rajpurkar, Anuj Pareek, Andrew~Y Ng, and Matthew~P Lungren.
\newblock Chexbert: combining automatic labelers and expert annotations for accurate radiology report labeling using bert.
\newblock \emph{arXiv preprint arXiv:2004.09167}, 2020.

\bibitem[Sotudeh et~al.(2020)Sotudeh, Goharian, and Filice]{sotudeh2020attend}
Sajad Sotudeh, Nazli Goharian, and Ross~W Filice.
\newblock Attend to medical ontologies: Content selection for clinical abstractive summarization.
\newblock \emph{arXiv preprint arXiv:2005.00163}, 2020.

\bibitem[Tang et~al.(2023)Tang, Peng, Wang, Ding, Durrett, and Rousseau]{tang2023less}
Liyan Tang, Yifan Peng, Yanshan Wang, Ying Ding, Greg Durrett, and Justin~F Rousseau.
\newblock Less likely brainstorming: Using language models to generate alternative hypotheses.
\newblock \emph{arXiv preprint arXiv:2305.19339}, 2023.

\bibitem[Toh et~al.(2012)Toh, Wei, Ng, Wan, Castillo, and Lin]{toh2012differentiation}
CH~Toh, K-C Wei, S-H Ng, Y-L Wan, M~Castillo, and C-P Lin.
\newblock Differentiation of tumefactive demyelinating lesions from high-grade gliomas with the use of diffusion tensor imaging.
\newblock \emph{American journal of neuroradiology}, 33\penalty0 (5):\penalty0 846--851, 2012.

\bibitem[Turpin et~al.(2023)Turpin, Michael, Perez, and Bowman]{turpin2023language}
Miles Turpin, Julian Michael, Ethan Perez, and Samuel~R Bowman.
\newblock Language models don't always say what they think: Unfaithful explanations in chain-of-thought prompting.
\newblock \emph{arXiv preprint arXiv:2305.04388}, 2023.

\bibitem[Van~Veen et~al.(2023)Van~Veen, Van~Uden, Attias, Pareek, Bluethgen, Polacin, Chiu, Delbrouck, Chaves, Langlotz, et~al.]{van2023radadapt}
Dave Van~Veen, Cara Van~Uden, Maayane Attias, Anuj Pareek, Christian Bluethgen, Malgorzata Polacin, Wah Chiu, Jean-Benoit Delbrouck, Juan Manuel~Zambrano Chaves, Curtis~P Langlotz, et~al.
\newblock Radadapt: Radiology report summarization via lightweight domain adaptation of large language models.
\newblock \emph{arXiv preprint arXiv:2305.01146}, 2023.

\bibitem[Wolf et~al.(2020)Wolf, Debut, Sanh, Chaumond, Delangue, Moi, Cistac, Rault, Louf, Funtowicz, Davison, Shleifer, von Platen, Ma, Jernite, Plu, Xu, Scao, Gugger, Drame, Lhoest, and Rush]{wolf2020huggingfaces}
Thomas Wolf, Lysandre Debut, Victor Sanh, Julien Chaumond, Clement Delangue, Anthony Moi, Pierric Cistac, Tim Rault, Rémi Louf, Morgan Funtowicz, Joe Davison, Sam Shleifer, Patrick von Platen, Clara Ma, Yacine Jernite, Julien Plu, Canwen Xu, Teven~Le Scao, Sylvain Gugger, Mariama Drame, Quentin Lhoest, and Alexander~M. Rush.
\newblock Huggingface's transformers: State-of-the-art natural language processing, 2020.

\bibitem[Yang et~al.(2022)Yang, Chen, PourNejatian, Shin, Smith, Parisien, Compas, Martin, Costa, Flores, et~al.]{yang2022large}
Xi~Yang, Aokun Chen, Nima PourNejatian, Hoo~Chang Shin, Kaleb~E Smith, Christopher Parisien, Colin Compas, Cheryl Martin, Anthony~B Costa, Mona~G Flores, et~al.
\newblock A large language model for electronic health records.
\newblock \emph{NPJ Digital Medicine}, 5\penalty0 (1):\penalty0 194, 2022.

\bibitem[Zhang et~al.(2023)Zhang, Press, Merrill, Liu, and Smith]{zhang2023language}
Muru Zhang, Ofir Press, William Merrill, Alisa Liu, and Noah~A Smith.
\newblock How language model hallucinations can snowball.
\newblock \emph{arXiv preprint arXiv:2305.13534}, 2023.

\bibitem[Zhang et~al.(2019)Zhang, Merck, Tsai, Manning, and Langlotz]{zhang2019optimizing}
Yuhao Zhang, Derek Merck, Emily~Bao Tsai, Christopher~D Manning, and Curtis~P Langlotz.
\newblock Optimizing the factual correctness of a summary: A study of summarizing radiology reports.
\newblock \emph{arXiv preprint arXiv:1911.02541}, 2019.

\end{thebibliography}

\appendix

\section{Prompting for Evidence}

\subsection{Single prompt}
\label{app:single-prompt}
For the single prompt setting (\S \ref{section:approach}), we followed \cite{mcinerney2020query}'s approach of using `future' ICD codes as the patient's diagnoses. To select a diagnosis that the patient did \textit{not} have, we randomly sampled an ICD code, excluding the patient's diagnoses. 

For Flan-T5, we use the following Chain of Thought (CoT) prompt :
\begin{quote}
Read the following clinical note of a patient: \texttt{[NOTE]}

Question: Is the patient at risk of \texttt{[DIAGNOSIS]}?

Answer: Let’s think step by step.
\end{quote}

\noindent Some examples of incorrect reasoning by Flan-T5:
\begin{enumerate}
    \item \textbf{Diagnosis:} Diabetes insipidus 
    
    \textit{The patient has a Swan-Ganz catheter in place. The Swan-Ganz catheter is a device that is used to measure the amount of blood circulating to the brain. The higher the blood flow to the brain, the higher the risk of diabetes insipidus. The answer: yes.}

    \item \textbf{Diagnosis:} Extradural and subdural abscess 
    
    \textit{The patient has asymmetrically distributed airspace opacities affecting the left lung more than the right with coexisting bilateral septal thickening is most suggestive of widespread pulmonary edema. Superimposed aspiration is also possible, particularly in the left lung. The heart remains enlarged. Small pleural effusions are new. The patient is at risk of extradural and subdural abscess.}
\end{enumerate}

For Mistral-Instruct, again we found that CoT prompting yielded very lengthy responses which were harder to parse. We thus used the following prompt:
\begin{quote}
Read the following clinical note of a patient: \texttt{[NOTE]}

Question: Is the patient at risk of \texttt{[DIAGNOSIS]}? Answer Yes or No and explain your answer. Be concise.

Answer:
\end{quote}

Some examples of incorrect reasoning by Mistral-Instruct:
\begin{enumerate}
    \item \textbf{Diagnosis:} Pneumocephalus 
    
    \textit{The patient has a history of taking Tylenol PR and has a cool bath, ice packs, and a fan to cool down, which may increase the risk of pneumocephalus.}

    \item \textbf{Diagnosis:} Stroke 
    
    \textit{The patient has an endotracheal tube in place, which can increase the risk of stroke if not properly positioned and secured.}
\end{enumerate}

\subsection{Few-shot prompting}
\label{app:few-shot} 
We provide additional details on our preliminary experiments with few-shot prompting. As done in \S \ref{app:single-prompt}, we followed \cite{mcinerney2020query}'s approach of using `future' ICD codes as the patient's diagnoses. To select a diagnosis that the patient did \textit{not} have, we randomly sampled an ICD code, excluding the patient's diagnoses.
We used the following prompt:
\begin{quote}
\noindent Read the following clinical note of a patient: \texttt{[RANDOM NOTE SNIPPET]}.

\noindent Answer step by step: can the patient possibly have cardioembolic strokes in the future?

\noindent Answer: There is no evidence. Final answer: No.
    
\noindent Read the following clinical note of a patient: patient stopped taking a blood thinning medication required for a heart valve due to side effects.

\noindent Answer step by step: can the patient possibly have cardioembolic strokes in the future?
    
\noindent Answer: The patient stopped taking a blood thinning medication required for a heart valve. The medication thins the blood and prevents blood clots. Blood clots can lead to strokes. Final answer: Yes.
    
\noindent Read the following clinical note of a patient: \texttt{[NOTE]}.
    
\noindent Answer step by step: based on the note, why is the patient at risk of \texttt{[DIAGNOSIS]}?
    
\noindent Answer: 
    
\end{quote}

We observed that with few-shot prompting the model surfaced evidence for almost every diagnosis that the patient did not have. For example, for a patient with \emph{`with g/j tube in place for gastroparesis'}, the model's output for the diagnosis, encephalitis, was \emph{`The patient has a jejunostomy tube in place. The jejunostomy tube can be pulled out. The jejunostomy tube can be pulled out of the body. The jejunostomy tube can be pulled out of the body and into the brain. Final answer: Yes'}. 

We suspect the prompt biases the model to support the query diagnosis which then makes the model generate incorrect explanations as evidence \citep{turpin2023language}. We also experimented with prompts such as \emph{`Extract evidence for \texttt{[DIAGNOSIS]}. Output N/A if no evidence exists'} but the model then mostly generated \emph{`N/A'}. 
Given these results, we carried the rest of the evaluation with the zero-shot prompting approach.

\begin{table*}
\centering
\footnotesize
\begin{tabular}{lp{12cm}}
\multicolumn{1}{l}{\textbf{Diagnosis}}&\multicolumn{1}{c}{\textbf{Risk Factors}}\\
\hline
pneumocephalus&head injury, skull fracture, neurosurgical procedures, sinus or mastoid surgery, meningitis, cerebrospinal fluid leak, barotrauma, diving or scuba diving accidents, iatrogenic causes, such as lumbar puncture or spinal anesthesia\\
stroke&hypertension, smoking, diabetes, obesity, sedentary lifestyle, high cholesterol levels, atrial fibrillation, family history of stroke, previous history of stroke, excessive alcohol consumption, drug abuse.\\
intracranial hemorrhage&hypertension, aneurysms, arteriovenous malformations, blood clotting disorders, trauma, drug abuse, liver disease, brain tumor, stroke, coagulopathy\\
brain tumor&progression	genetics, exposure to ionizing radiation, family history of brain tumors, certain hereditary conditions, weakened immune system, previous history of brain tumor.\\
intracranial hypotension&obesity, connective tissue disorders, previous spinal or cranial surgery, leaking cerebrospinal fluid, spinal epidural anesthesia, lumbar puncture or spinal tap\\

\end{tabular}
\caption{Examples of risk factors provided by GPT-3.5}
\label{tab:risk_factors}
\end{table*}

\begin{table*}
    \centering
    \footnotesize
    \begin{tabular}{p{3.45cm}p{4cm}p{8cm}} 
    \textbf{Diagnosis}&\textbf{Evidence}&\textbf{Explanation}\\
    \hline
    intracranial hemorrhage&patient had multiple cardiac surgeries&Multiple cardiac surgeries may suggest anticoagulation or underlying cardiac dysfunction which could in turn predispose the patient to intracranial hemorrhage.\\ 
    \hline
    intracranial hypotension&The patient has a ventriculoperitoneal shunt.&A ventriculoperitoneal shunt (VPS) is a surgical device used to relieve intracranial pressure by draining excessive cerebrospinal fluid. Having a VPS catheter may increase the risk of intracranial hypotension due to over drainage. \\ 
    \hline
    craniopharyngioma&s/p resection X2, s/p VPS and panhypopitutiarism with second resection&Panhypopituitarism and the fact that something was removed through surgery suggests there was a tumor involving the sella which may or may not have been craniopharyngioma.
    \end{tabular}
    \caption{Examples of weakly correlated evidence surfaced by the model for different diagnosis queries. All have plausible but somewhat removed (or weak) connections. }
    \label{tab:weak_corr_examples}
\end{table*}

\section{Automatic Evaluation}
\label{app:auto_eval}
Our proposed LLM-based automatic evaluation (Section \ref{sec:auto-eval}) consists of three steps, each realized as a single prompt. We use a one-shot prompt for the first step and zero-shot prompts for the subsequent steps, as shown below.

\begin{enumerate}
    \item Extract risk factors from the evidence.

    \begin{quote}
    Read the following statement: The patient is at risk of intracranial hemorrhage due to presence of an endotracheal tube (ETT) in the patient's trachea which may increase the risk of complications such as aspiration and airway obstruction.
    
    Question: Extract the risk factors from the statement as a list. Be concise.
    
    Answer: 1. presence of endotracheal tube (ETT) in the trachea.
    
    Read the following statement: \texttt{[EVIDENCE]}
    
    Question: Extract the risk factors from the statement as a list. Be concise.
    
    Answer: "
    \end{quote}

    \item Verify the presence of each risk factor in the note.

    \begin{quote}
        Read the following clinical note of a patient: \texttt{[NOTE]}
        
        Question: Does the patient have \texttt{[RISK FACTOR]}? Answer Yes or No. 
    \end{quote}
    
    \item  Validate if each present risk factor is a valid risk factor of query diagnosis.
    
    \begin{quote}
        Is \texttt{[RISK FACTOR]} a risk factor of \texttt{[DIAGNOSIS]}? Choice: -Yes -No. Be concise.
        
        Answer: 
    \end{quote}
\end{enumerate}

We used the following prompts for signs:
\begin{enumerate}
    \item Extract signs from the evidence.

    \begin{quote}
    Read the following statement: A patient may have intracranial hemorrhage because of the following report - Large left subdural hematoma, extensive subarachnoid hemorrhage, right temporal effacement, left uncal herniation, and effacement of the sulci.
    
    Question: Extract the signs from the statement as a list. Be concise.
    
    Answer: 1. Large left subdural hematoma 2. Extensive subarachnoid hemorrhage 3. Right temporal effacement 4. Left uncal herniation 5. Effacement of the sulci
    
    Read the following statement: A patient may have \texttt{[DIAGNOSIS]} because of the following report - \texttt{[EVIDENCE]}.
    
    Question: Extract the signs from the statement as a list. Be concise.
    
    Answer: "
    \end{quote}

    \item Verify the presence of each sign in the note.

    \begin{quote}
        Read the following clinical note of a patient: \texttt{[NOTE]}
        
        Question: Does the patient have \texttt{[SIGN]}? Answer Yes or No. 
    \end{quote}
    
    \item  Validate if each present sign is a valid sign of query diagnosis.
    
    \begin{quote}
        A patient is showing the following sign: \texttt{[SIGN]}.
        
        Question: Can the sign indicate \texttt{[DIAGNOSIS]}? Choice: -Yes -No. Be concise.
        
        Answer: 
    \end{quote}
\end{enumerate}

\begin{figure*}
\centering
    \includegraphics[scale=.4]{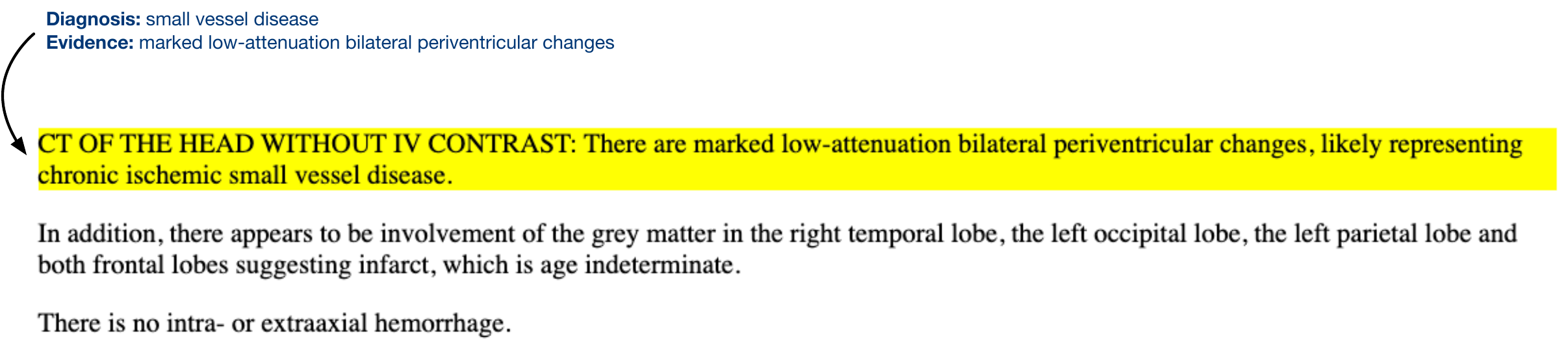}
    \caption{Screenshot of the evaluation interface showing highlighted evidence.}
    \label{fig:interface}
\end{figure*}

\section{Binary decision recall}
\label{app:recall}

Recall that we first ask the LLM whether a note indicates that the corresponding patient is at risk for, or has, a given query diagnosis. 
The precision of this LLM inference 
is implicitly measured by the assessment of generated evidence; if the patient does not have (is not at risk for) a condition, generated evidence will necessarily be irrelevant. 
But this does not capture model recall, i.e., recognizing when a patient indeed has (is at risk of) a condition. 

To also estimate model \emph{recall}, we sampled $20$ patients from BWH and followed prior work \citep{mcinerney2020query} in our  evaluation. 
Specifically, we asked radiologists to browse reports from up to one year following a reference radiology report 
and tag relevant diagnoses; these constitute ``future'' diagnoses with respect to the reference report. 
Radiologists then flagged past notes containing supporting evidence for these diagnoses. 
Of the $200$ notes marked as containing evidence, Mistral-Instruct, Flan-T5, and CBERT had a recall of $58.2$, $70.0$, and $80.4$ respectively.

\section{Likely Indicators}
\label{app:likely_indicators}
For the \textit{likely indicators} in \S \ref{section:scaling-eval}, we used `likely represent', 'concerning for', and `diagnosis include'. We did not consider diagnoses such as `old infarction', which came up often for `likely represent'. An infarction can be myocardial or cerebral. Since our dataset comprises of radiology reports concerning brain scans, we added 'cerebral' as prefix to `infarction' to ensure specificity. Similarly, we added `brain' as a prefix to `metastasis'.

\begin{table}
    \centering
    \small
    \begin{tabular}{lccc}
    Model&\% instances&\# evidence &\# risks\\
    &with evidence&&(signs)\\
    \hline
    Flan-T5\\
    \hspace{0.8em}MIMIC&$91.0$&$1,077$&$2,817$\\
    \hspace{0.8em}BWH&$88.0$&$701$&$2,027$\\
    Mistral-Instruct\\
    \hspace{0.8em}MIMIC&$84.0$&$968$&$2,894$\\
    \hspace{0.8em}BWH&$90.0$&$614$&$1,799$\\
    CBERT\\
    \hspace{0.8em}MIMIC&$100.0$&$2,000$&$7,467$\\
    \hspace{0.8em}BWH&$100.0$&$1,000$&$3,336$\\
    \hline
    \end{tabular}
    \caption{Data statistics of large-scale evaluation performed in \S  \ref{section:scaling-eval}. We evaluated $100$ and $50$ instances from MIMIC and BWH datasets respectively.}
    \label{tab:scaled-up-eval-stats}
\end{table}

\section{Implementation Details}
We used the \texttt{HuggingFace} \citep{wolf2020huggingfaces} library to run inference using Mistral-Instruct (7B), Flan-T5 XXL (11B) and ClinicalBERT ($110$ million parameters). We split notes into sentences using the \texttt{spaCy} (en\_core\_web\_sm) library \citep{spacy2}. We processed notes in chunks of size $750$ tokens (including the prompt text) for Flan-T5 and Mistral-Instruct. We used a single NVIDIA Tesla V100 ($32$G) GPU.

\end{document}